\pdfoutput=1

\documentclass[11pt]{article}

\usepackage[]{emnlp2021}

\usepackage{times}
\usepackage{latexsym}

\usepackage[T1]{fontenc}

\usepackage[utf8]{inputenc}

\usepackage{microtype}

\usepackage{times}
\usepackage{latexsym}
\usepackage{float}
\usepackage{comment}
\usepackage{amsmath,amsfonts,amssymb,amsthm}
\usepackage{mathtools}
\usepackage{linguex}

\usepackage[makeroom]{cancel}

\usepackage{graphicx}

\usepackage{caption}
\usepackage{subcaption}

\usepackage{courier}

\usepackage{bm}
\usepackage{adjustbox}
\usepackage{flushend}

\newcommand{\norm}[1]{\left\lVert#1\right\rVert}

\DeclareMathOperator*{\argmax}{arg\,max}

\title{Attention Weights in Transformer NMT Fail Aligning Words Between Sequences but Largely Explain Model Predictions}

\author{Javier Ferrando \and Marta R. Costa-jussà \\
         TALP Research Center, Universitat Politècnica de Catalunya, Barcelona \\
         \texttt{\{javier.ferrando.monsonis,marta.ruiz\}@upc.edu}}

\begin{document}
\maketitle
\begin{abstract}

This work proposes an extensive analysis of the Transformer architecture in the Neural Machine Translation (NMT) setting. Focusing on the encoder-decoder attention mechanism, we prove that attention weights systematically make alignment errors by relying mainly on uninformative tokens from the source sequence. However, we observe that NMT models assign attention to these tokens to regulate the contribution in the prediction of the two contexts, the source and the prefix of the target sequence. We provide evidence about the influence of wrong alignments on the model behavior, demonstrating that the encoder-decoder attention mechanism is well suited as an interpretability method for NMT. Finally, based on our analysis, we propose methods that largely reduce the word alignment error rate compared to standard induced alignments from attention weights.
\end{abstract}

\section{Introduction}

Recently, Transformer-based models \cite{NIPS2017_3f5ee243} have allowed huge improvements in performance across multiple NLP tasks. The inclusion of this architecture has led the field of NLP to investigate the inner workings of this architecture in several tasks. One of its core components, the attention mechanism, which provides a distribution of scores over the input tokens, has been
often presented as showing the relative importance of the inputs. Some works have criticized the use of attention weights as model explanations \cite{jain-wallace-2019-attention,serrano-smith-2019-attention,pruthi-etal-2020-learning}, demonstrating that attention weights distributions can be modified without affecting the final prediction. However, these studies have mainly analyzed encoder-only or decoder-only architectures like BERT \cite{devlin-etal-2019-bert} or GPT-2 \cite{Radford2019LanguageMA}, which are based on self-attention mechanisms.

Nonetheless, NMT models use the encoder-decoder Transformer architecture, which adds the encoder-decoder attention mechanism, in charge of distributing the information flow from the encoder representations of the source input tokens into the decoder. \cite{voita-etal-2019-analyzing} analyze the effect of pruning different attention heads in a Transformer NMT model and conclude that the encoder-decoder attention mechanism is the most critical one. \cite{raganato-etal-2020-fixed} show that encoder self-attention weights can be interchanged by predefined non-learnable patterns without hindering the translation performance. These results provide evidence about the relevance of the encoder-decoder attention mechanism on NMT, which we believe needs further investigation. In this work we analyze the encoder-decoder attention weights and shed light on their impact on the decoder representations and final predictions, showing how alignment errors can also give information about the model's decision-making process.

Research in NMT interpretability has mainly focused on understanding source words importance when predicting a target word. The word alignment task \cite{och-ney-2003-systematic} has served to compare explanation methods against human-annotated source-target word alignments. Encoder-decoder attention weights have been used to provide source-target word alignments \cite{Zenkel_2019,garg_jointly_2019}, but its low performance has made researchers sceptical about its use as an interpretable method \cite{li-etal-2019-word}. 
An important issue when relying on word alignment task is that it ignores the words that are predicted based on the target prefix, i.e what the model has previously translated. An extreme example of the impact of the target prefix on the prediction occurs during 'hallucinations' \cite{lee2019hallucinations,berard-etal-2019-naver,voita_analyzing_2020,raunak2021curious}. Although some studies have analyzed the relative contribution of the target prefix context in a model's prediction \cite{li-etal-2019-word,voita_analyzing_2020}, the way NMT models decide in which proportion to use both sequences remains unexplored.

In Section \ref{sec:ana}, we propose a simple method to measure the relative contribution of the source and the target prefix by perturbing input embeddings, we also extend the gradient-based method towards the target prefix token embeddings to obtain saliency scores from the prefix words. Both methods serve us in Section \ref{sec:Analysis} to understand the attention weights generated in the encoder-decoder modules and their relationship with the model predictions. Lastly, in Section \ref{sec:improving_alignment}, we propose two methods to improve the alignment error extracted from the model in accordance with our results analysis.

\section{Background}

In this section, we briefly introduce the existing methodologies that we use in our work: the Transformer and the methods used to induce alignment.  

\subsection{Transformers in NMT}
Given a source sequence $\mathbf{x} = \{x_1,\cdots,x_{|\mathbf{x}|}\}$\footnote{Along this work we use $x$ to represent elements (scalars/words/tokens), $\bm{x}$ vectors, $\mathbf{x}$ sequences and $\mathbf{X}$ matrices.} and a target sequence $\mathbf{y}=\{y_1,\cdots,y_{|\mathbf{y}|}\}$ an NMT system models the probability:
\begin{align*}
P(\mathbf{y}|\mathbf{x}) &= \prod_{t=1}^{|\mathbf{y}|} P(y_t|\mathbf{y_{<t}},\mathbf{x})
\end{align*}

where $\mathbf{y_{<t}}=\{y_0,\cdots,y_{|t-1|}\}$ represent the prefix of $y_{t}$, and $x_{|\mathbf{x}|} = y_{0} = y_{|\mathbf{y}|} = \texttt{$\langle \text{/s} \rangle$}$, which represents a special token used to denote the beginning and end of sentence. The Transformer architecture is composed by a stack of encoder layers and decoder layers. The encoder generates a contextualized sequence of representations $\mathbf{e} = \{\bm{e}_1,\cdots,\bm{e}_{|\mathbf{x}|}\}$ of the source sentence while the decoder, at each time step $t$, uses both the encoder output and the token representation $\bm{s}_{t}^{l-1}$ of the previous layer $l$ to compute the final probability distribution over the target vocabulary.

In terms of the model's input and output, we consider $\bm{x}_{j}$ and $\bm{y}_{i}$ representing the embeddings of each token from the source and target prefix respectively. So, we can write the conditional probability modeled by the network at each time step as:
\begin{gather*}\label{eq_p}
P(y_t|	\{ \bm{y}_{0},\cdots,\bm{y}_{t-1}\},\{ \bm{x}_{1},\cdots,\bm{x}_{|\mathbf{x}|}\})
\end{gather*}

\begin{figure}[h!]
	\begin{centering}
	\includegraphics[width=0.5\textwidth]{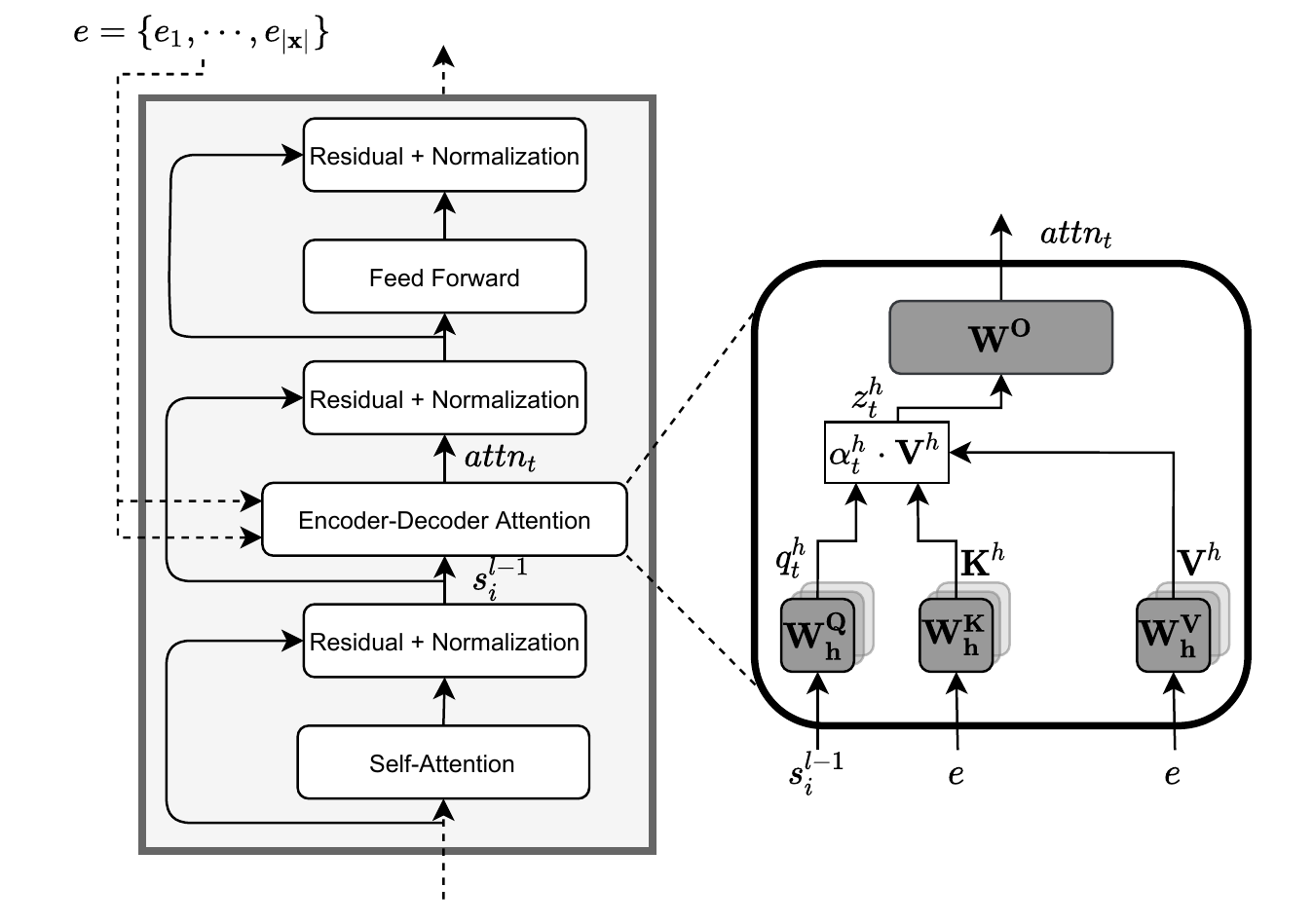}
	\caption{Decoder Layer with the Encoder-Decoder Attention module expanded.}
	\label{fig:enc_dec}
	\end{centering}
\end{figure}

The encoder and decoder representations are merged in the multi-head encoder-decoder attention mechanism (Figure \ref{fig:enc_dec}). For each head, the encoder embeddings are projected to keys and values. Formally, $\mathbf{V}^{h} \in \mathbb{R}^{|\mathbf{x}| \times d_{v}}$ is the value matrix and $\mathbf{K}^{h} \in \mathbb{R}^{|\mathbf{x}| \times d_{k}}$ is the key matrix, where $d_{v}$ and $d_{k}$ refers to the dimension of the values and keys vectors. The decoder representation of the output token $\bm{s}_{t}^{l-1}$ is projected to a query vector of dimension $d_{q}$, $\bm{q}^{h}_{t} \in \mathbb{R}^{d_{q}}$. The output of each attention head is obtained by:
\begin{equation}\label{eq:z_t}
\bm{z}^{h}_{t} = \sum_{j=1}^{|\mathbf{x}|}\bm{\alpha}^{h}_{t,j}\bm{v}^{h}_{j}
\end{equation}

Where:
\[
\bm{\alpha}^{h}_{t}=\text{softmax} \left(\frac{\bm{q}^{h}_{t}\mathbf{K}^\top}{\sqrt{d_{k}}}\right)
\]

$\bm{\alpha}_{t}^{h}$ refers to the vector of attention scores at decoding step $t$, which is often presented as a matrix (attention matrix) made of a stack of $\bm{\alpha}^{h}_{t}$, for every time step.
This process is repeated simultaneously in multiple heads. Each head computes a $\bm{z}^{h}_{t}$ representation, and are concatenated before projecting by $\mathbf{W_{h}^{O}}$ to obtain $\bm{attn}_t$.

\subsection{Attention Weights to Induce Word Alignment}\label{sec:attn_alignments}

Attention weights $\alpha^{h}_{t,j}$ from the encoder-decoder attention modules represent the similarity between $\bm{h_{j}}$ and $\bm{s}_{t}^{l-1}$ and have been commonly presented as a baseline to extract word alignments from words $x_{j}$ and $y_{t}$. Attention vectors $\bm{\alpha_{t}}^h$ represent a probability distributions over all source tokens $\mathbf{x}$. A classical approach to obtain final alignments has been to compute the average over all heads \cite{garg_jointly_2019} in each layer and selecting the source word that yields the maximum score:
\[
  \mathbf{A}_{t,j} = \left\{
     \begin{array}{@{}l@{\thinspace}l}
       1 \quad j= \argmax_{j'}\frac{1}{H}\sum_{h=1}^{H} \alpha_{t,j'}^{h} \\
       0 \quad \text{else}\\
     \end{array}
   \right.
\]

\cite{Zenkel_2019,li-etal-2019-word,garg_jointly_2019} showed alignments induced from attention weights are noisy, although they realize that some layers seem to generate better $(x_{j},y_{t})$ alignments, especially the last layers of the Transformer.

An issue regarding the use of this method to interpret the model predictions is that the ground truth target word $y_{t}$ may differ from the actual model prediction $y'_{t}$. In these cases, $(x_{j}$,$y'_{t})$ alignments can not be compared with $(x_{j}$,$y_{t})$ gold alignments, showing limitations about its use as an interpretability method.

\paragraph{Aligments from the decoder input.}
A technique that solves the aforementioned issue consists of inducing alignments by comparing $\mathbf{x}$ with the input of the decoder $y_{i}$ \cite{kobayashi-etal-2020-attention,chen-etal-2020-accurate} (in force decoding setting $y_{i} = y_{t-1}$). So, since the ground truth target sequence is used as input in the decoder, alignments $\mathbf{A}_{i,j}$ in this setting represent the same information as gold alignments. Attention modules from the initial layers tend to extract better alignments from the input of the decoder, while alignments from the decoder output are better extracted from the final layers. Although results show that decoder input provides lower alignment error rates, it shows how similar to $\bm{e}_{j}$ the model is able to generate representations of the decoder input, losing explanation power about the influence of source tokens into the model output. Therefore, we use $\mathbf{A}_{t,j}$ in our analysis in Section \ref{sec:Analysis}. An extension of the use of attention weights to induce alignments is presented in \cite{kobayashi-etal-2020-attention}, where it is also considered the norm of the vectors projected by the linear layers inside the attention modules.

\subsection{Other Methods}
\paragraph{Model-agnostic methods.} Several methods for inducing alignments have been proposed that work regardless of the chosen architecture. Gradient-based methods such as gradient $\times$ input \cite{ding-etal-2019-saliency} or Integrated Gradients \cite{he-etal-2019-towards} have been used to obtain saliency values from the source words as a measure of source word importance. Erasure methods have also been applied to NMT \cite{li-etal-2019-word}, which consist of techniques to measure the relevance of each input token by evaluating the changes in the output probability of the model after removing it from the input of the network \cite{DBLP:conf/iclr/ZintgrafCAW17} or eliminating the connection via dropout \cite{JMLR:v15:srivastava14a}.
\paragraph{Methods to improve alignments.} Other works propose methods to improve word alignment extracted from the Transformer. \cite{li-etal-2019-word} use an explicit alignment model \cite{liu-etal-2005-log,taskar-etal-2005-discriminative} consisting of optimizing a parameter matrix to reduce the alignment distance with respect to a reference. \cite{Zenkel_2019} adds an alignment module attending encoder representations. \cite{garg_jointly_2019} propose to supervise an attention head with GIZA++ \cite{Brown1994} alignments. Although they improve alignment performance, these methods introduce external trainable parameters or alignments references, which makes these techniques lose interest regarding interpretability of the model.

\section{Proposed Methods for Analysis}
\label{sec:ana}

In this section, we introduce two simple methods for measuring the contributions of each source sequence to a model prediction and extend the gradient-based analysis to understand dependency relationships between target prefix words.

\begin{figure}[H]
	\begin{centering}
	\includegraphics[width=0.5\textwidth]{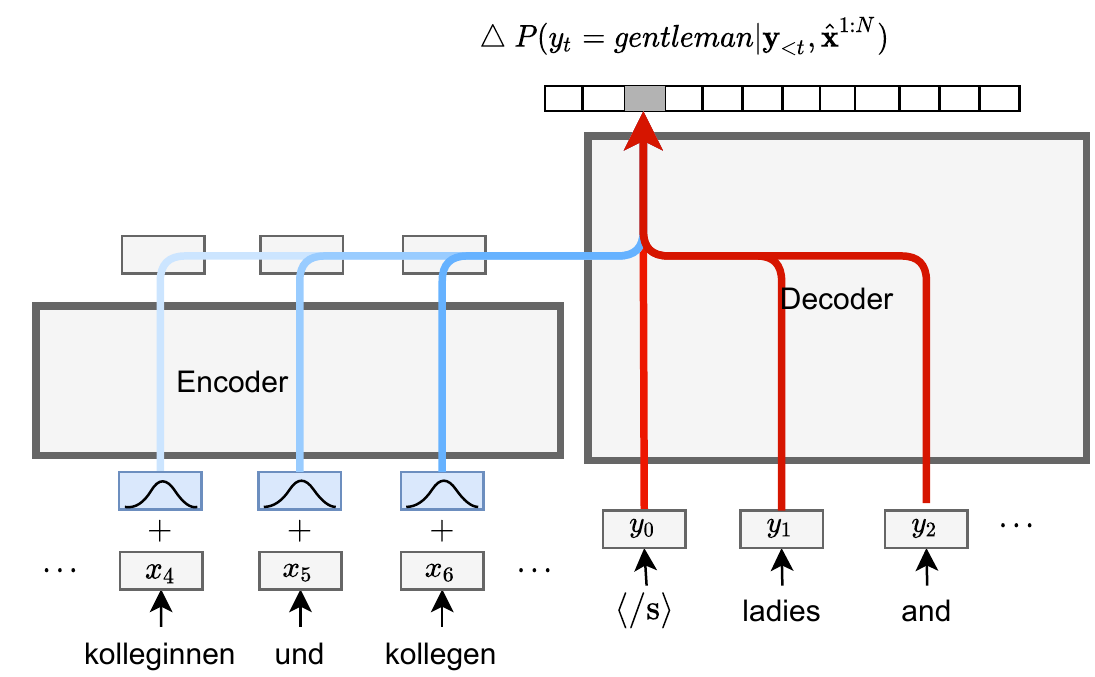}
	\caption{Contribution from the source input sequence to the final prediction by perturbing source token embeddings.}
	\label{fig:noisy_input}
	\end{centering}
\end{figure}

\subsection{Contributions by Input Perturbation}\label{sec:source_contributions}
We propose separately perturbing source and prefix embeddings \cite{DBLP:journals/corr/SmilkovTKVW17} to get the marginal contributions of each sequence to the final prediction. For each embedding we compute $N$ random samples around their neighborhood:
\[
\hat{\bm{x}}_{j} = \bm{x}_{j} + \mathcal{N}(0,\sigma_{\bm{x}_j}^2)
\]
Since input embeddings differ in their length, we adapt the noise level to each token embedding as a proportion ($\lambda$) of its euclidean norm\footnote{In this work we use $\lambda =1\%$}:
\[
\sigma_{\bm{x}_j} = \norm{\bm{x}_j}\cdot \lambda
\]
By adding noise to each embedding in the sequence we get the perturbed sequence of embeddings $\hat{\mathbf{x}}$.
So, for each prediction $y_t$ we can compute the source contribution $C_S(y_t)$ by measuring how large is the variation of the output probability when feeding the network with $N$ noisy sequence samples. To get the marginal effect of one sequence, we keep the other with the original embeddings.
\begin{figure}[H]
	\begin{centering}
	\includegraphics[width=0.49\textwidth]{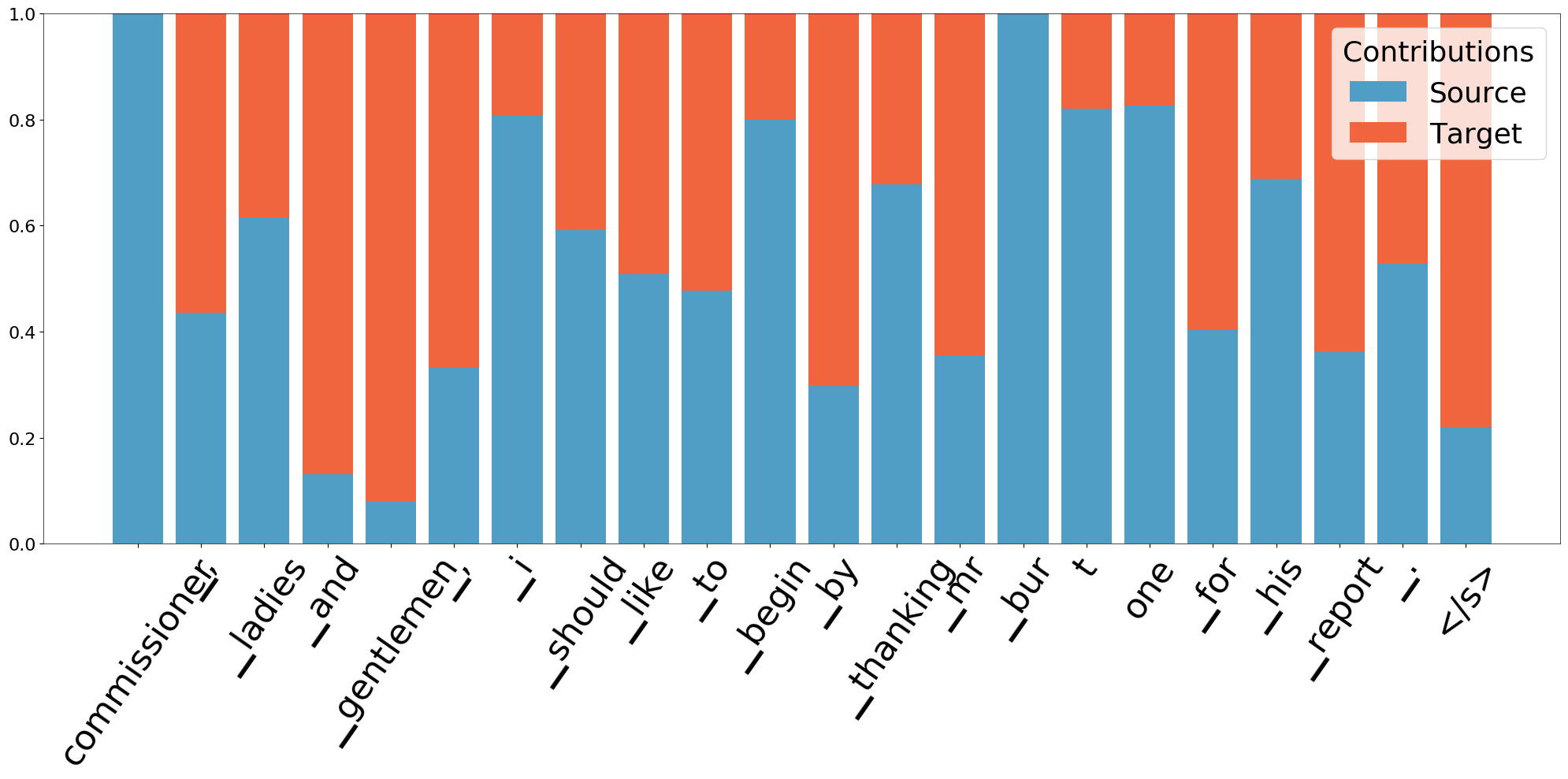}
	\caption{Source $C_S(y_t)$ and target prefix $C_T(y_t)$ contributions for reference model output.}
	\label{fig:source_target_contributions}
	\end{centering}
\end{figure}
\begin{gather*} 
C_S(y_t) = \bigtriangleup P(y_{t}|\mathbf{y},\hat{\mathbf{x}}^{1:N})\\
=\frac{1}{N}\sum_{n=1}^{N}(P(y_t|\mathbf{y}_{<t},\hat{\mathbf{x}}^{n}) - \bar{P}(y_t|\mathbf{y}_{<t},\hat{\mathbf{x}}^{1:N}))^2
\end{gather*}
where $\bar{P}$ refers to the mean of the observed output probabilities and $\hat{\mathbf{x}}^{n}$ to $n$-th source sequence with added noise. Similarly, we get the target prefix contribution $C_T(y_t)$ perturbing prefix embeddings:
\[
\hat{\bm{y}}_{i} = \bm{y}_{i} + \mathcal{N}(0,\sigma_{\bm{y}_{i}}^2)
\]
and then, computing the variance of the output probability across $N$ sequences of noisy prefix embeddings, keeping untouched the original source token embeddings:
\begin{gather*} 
C_T(y_t) = \bigtriangleup P(y_{t}|\hat{\mathbf{y}}_{<t}^{1:N},\mathbf{x})\\
= \frac{1}{N}\sum_{n=1}^{N}(P(y_t|\hat{\mathbf{y}}_{<t}^n,\mathbf{x}) - \bar{P}(y_t|\hat{\mathbf{y}}_{<t}^{1:N},\mathbf{x}))^2
\end{gather*}
\subsection{Saliency of Target Sequences Words}
Any model $f(\bm{x})$ can be linearly approximated locally by its first-order Taylor expansion at a point $\hat{\bm{x}}$:
\[
f(\hat{\bm{x}}) \approx f(\bm{x}) + \nabla_{\bm{x}} f(\bm{x}) \cdot (\hat{\bm{x}} - \bm{x})
\]
Rearranging terms we get:
\[
f(\bm{x}) \approx f(\hat{\bm{x}}) + \nabla_{\bm{x}} f(\bm{x}) \cdot (\bm{x}-\hat{\bm{x}})
\]
Making $\hat{\bm{x}}$ a zero vector, we arrive to:
\[
f(\bm{x}) \approx \nabla_{\bm{x}} f(\bm{x}) \cdot \bm{x}
\]
With this approximation, $\nabla_{\bm{x}} f(\bm{x})$ can be interpreted as coefficients that measure the impact of $\bm{x}$ in the output. In NLP \cite{li-etal-2016-visualizing} propose the use of word embeddings as input features from which to calculate saliency scores. 
In the NMT setting, current methods \cite{ding-etal-2019-saliency} extract saliency scores of the input source tokens by computing the gradient with respect to source embeddings $\bm{x}_{i}$.

Nevertheless, the Transformer model deals with two different sequences of inputs ($\mathbf{x}$ and $\mathbf{y_{<t}}$), $f(\bm{x}) = P(y_t|\mathbf{y_{<t}},\mathbf{x})$. So, analyzing only the saliency of the source sequence embeddings might lead to an incomplete analysis. To have a full understanding of the influences of each input word on the model prediction we propose to extend the SmoothGrad method \cite{DBLP:journals/corr/SmilkovTKVW17} to also consider the gradients w.r.t the target prefix embeddings.
We compute the target prefix saliencies by averaging the gradients over $N$ noisy examples, as detailed in Section \ref{sec:source_contributions}:
\[
\psi(y_{i},y_{t}) = \frac{1}{N}\sum_{n=1}^{N}\norm{\nabla_{\bm{y}_{i}} P(y_t|\hat{\mathbf{y}}_{<t}^{n},\mathbf{x})}
\]
\begin{figure*}[ht!]
\centering
\begin{minipage}[b]{\textwidth}
\includegraphics[width=1\textwidth]{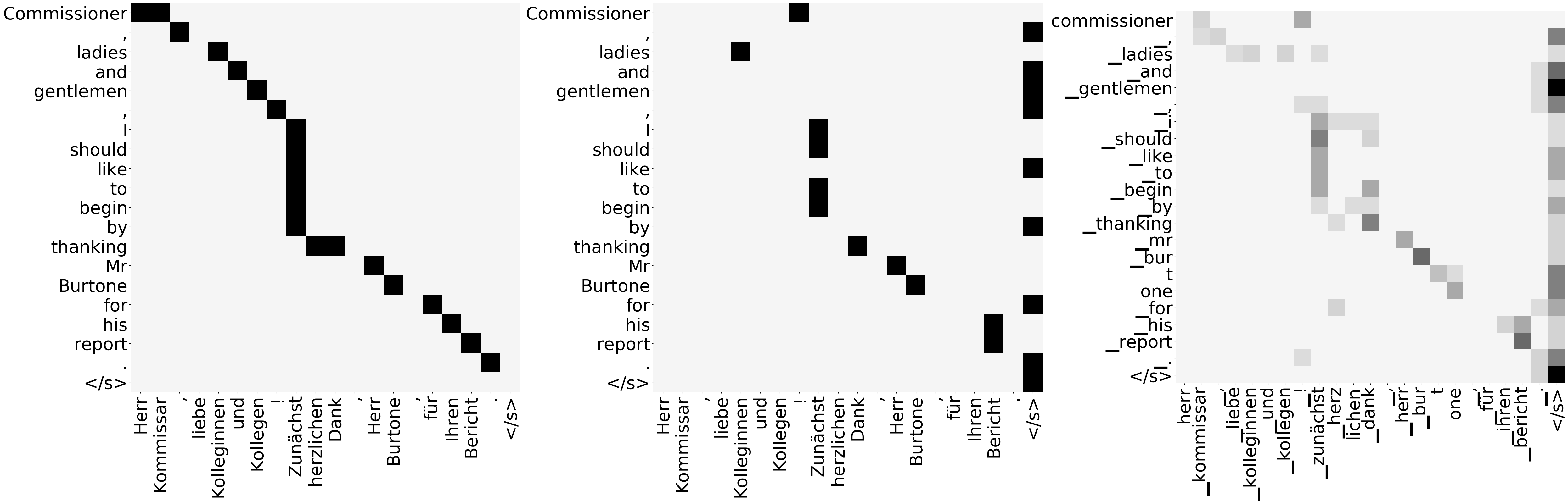}
\end{minipage}\hfill
\caption{From left to right: gold alignments, hard alignments $\mathbf{A}_{t,j}$ and soft alignments given by the average attention weights over all heads in Layer 5.}
\label{fig:alignments}
\end{figure*}
\section{Experimental Setup}

As follows, we detail the model and datasets used in our experiments. We decide to choose this experimental framework to compare and further explain previous works \cite{ding-etal-2019-saliency,Zenkel_2019,kobayashi-etal-2020-attention}. We follow the same procedure as these past works, we train the Transformer model for the German-English translation task. Specifically, we use Europarl v7 corpus\footnote{\url{http://www.statmt.org/europarl/v7}} which consists on 1.9M sentence pairs.
We use the gold alignment dataset\footnote{\url{https://www- i6.informatik.rwth-aachen.de/goldAlignment/}} \cite{Vilar2006AERDW}  which contains 508 sentence pairs. The Transformer used in this work \footnote{\texttt{transformer\_iwslt\_de\_en}} is implemented in fairseq \cite{ott-etal-2019-fairseq} and contains 6 layers with 4 attention heads each. We apply Byte Pair Encoding (BPE) \cite{sennrich-etal-2016-neural} with 10k merging operations. As \cite{ding-etal-2019-saliency} and \cite{kobayashi-etal-2020-attention} we use the last 1000 samples of the training data as the development data.

\section{Analysis}\label{sec:Analysis}
In this section, and for the sake of clarity, we convey our analysis on a single example while quantifying how our findings generalize to the entire test set. We use the following source example:

\ex.\label{ex:source} \textit{Herr Kommissar, liebe kolleginnen und kollegen! Zunächst herzlichen Dank, Herr Burtone, für Ihren Bericht.}

for which the model prediction is:

\ex.\label{ex:prediction} \textit{Mr, ladies and gentlemen, first would like to start by thanking Mr Burtone for his report.}

and with its reference\footnote{We refer to the words that are predicted with the highest probability as the model prediction $y'_{t}$, and the reference (ground truth) words as the ground truth or reference model output $y_{t}.$ 
}:

\ex.\label{ex:target} \textit{Commissioner, ladies and gentlemen, I should like to begin by thanking Mr Burtone for his report.}

\subsection{Categorization of Word Alignment Errors}
\label{sec:categ}

Figure \ref{fig:alignments} (Middle) shows ($x_{j},y_{t}$) alignments $\mathbf{A}_{t,j}$ extracted from the best layer (\S \ref{sec:attn_alignments}) for the example. Figure \ref{fig:alignments} (Right) depicts the average attention weight matrix across all heads in the best layer (soft alignments), from which some information can be recovered. When comparing hard with gold alignments (Figure \ref{fig:alignments} (Left)), some errors are clearly observed, with a large number of target tokens aligning to finalizing tokens. Hereinafter, finalizing tokens correspond to the special token used to indicate end of sentence (\texttt{$\langle \text{/s} \rangle$}) and the final punctuation mark (\texttt{\_.}), while the rest of tokens will be referred to as \textit{standard tokens}.

We categorize alignment errors occurring in weight attention matrices as:

\begin{enumerate}

    \item Functional/content words aligning to finalizing tokens.
    \item Words with non-direct translation aligning to finalizing tokens.
    \item Last tokens of a split word (divided into multiple subwords) aligning to finalizing tokens.
    \item Functional words aligning to the next content word token.
    \item Words aligning to other standard tokens.
    
\end{enumerate}

Our analysis focuses on finding explanations to the errors inside categories 1-3, which account for 60.6\% and 38.7\% of the total errors in the best layer in the ($x_{j},y_{t}$) and ($x_{j},y_{i}$) alignment settings respectively.


\subsection{Encoder-Decoder Attention Module decides Source-Target Contributions}
\label{sec:contrib}

\begin{figure}[H]
	\begin{centering}
	\includegraphics[width=0.48\textwidth]{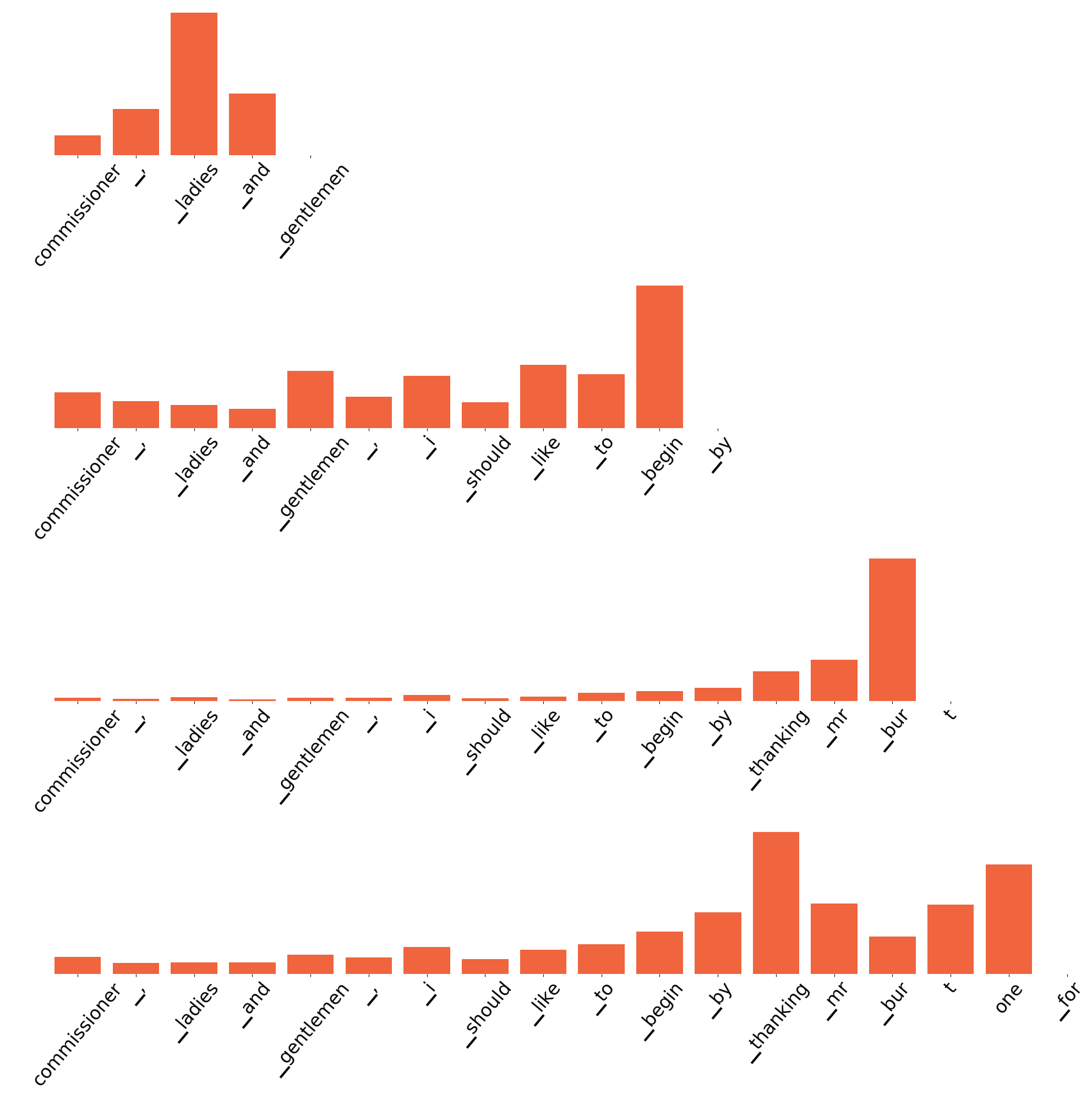}
	\caption{Saliency scores $\psi(\mathbf{y}_{<t},y_{t})$ for the reference model output (from top to bottom): \texttt{\_gentlemen}, \texttt{\_by}, \texttt{\_t} and \texttt{\_for}.}
	\label{fig:prefix_grad}
	\end{centering}
\end{figure}

From source-target contributions of the reference model output (Figure \ref{fig:source_target_contributions}), we observe that the target prefix largely contributes when predicting \textit{gentlemen}, and it also receives large saliency scores (Figure \ref{fig:prefix_grad}) from \textit{ladies}. This matches the human intuition about how these words are naturally generated from the context.

Similarly, a non-common word such as \textit{Burtone}, which gets tokenized into \texttt{\_bur}, \texttt{t} and \texttt{one} gets source-target contributions that also match human intuition. The first token \texttt{\_bur} is predicted by relying almost only on the source sequence. However, following tokens, although they heavily rely on the source, get information about the previous tokens. In this case, \texttt{\_bur} gives a high saliency value when predicting \texttt{t}. We can also observe that the word \textit{by} is mainly predicted using target prefix, and gets the highest saliency score from \textit{begin}. Another observation is that \textit{thanking} highly influences the prediction of \textit{for}. These examples have in common both large dependency on the target prefix and large attention values towards finalizing tokens.

\begin{figure}[h!]
	\begin{centering}
	\includegraphics[width=0.48\textwidth]{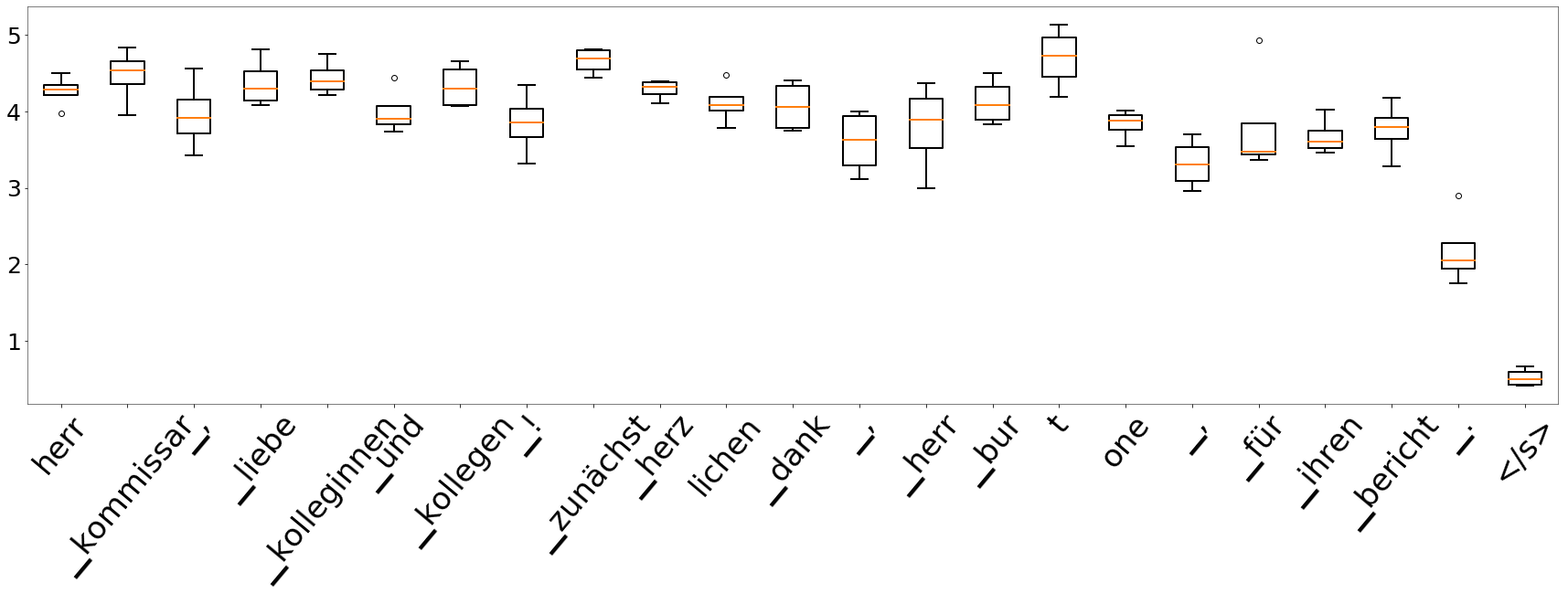}
	\caption{$\norm{v_{j}^{h}}$ computed in every attention head. \texttt{$\langle \text{/s} \rangle$} has almost zero norm for every head.}
	\label{fig:attn_bypass}
	\end{centering}
\end{figure}

\begin{figure}[h!]
\centering
\includegraphics[width=0.48\textwidth]{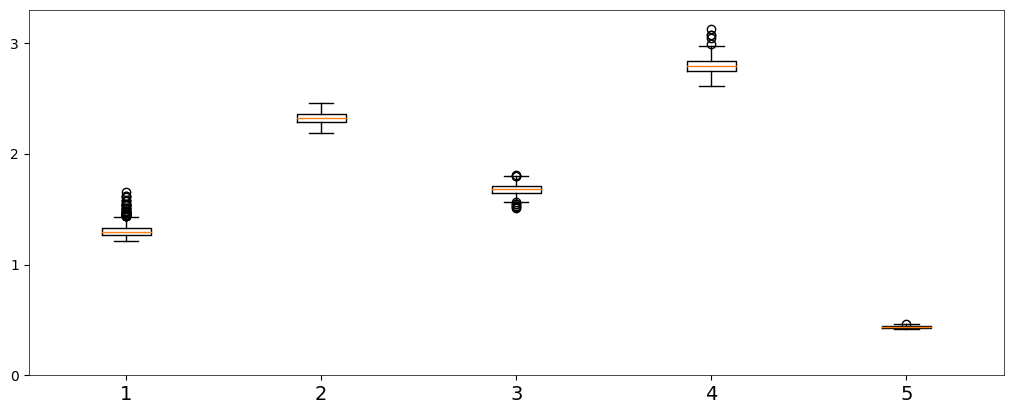}
\includegraphics[width=0.48\textwidth]{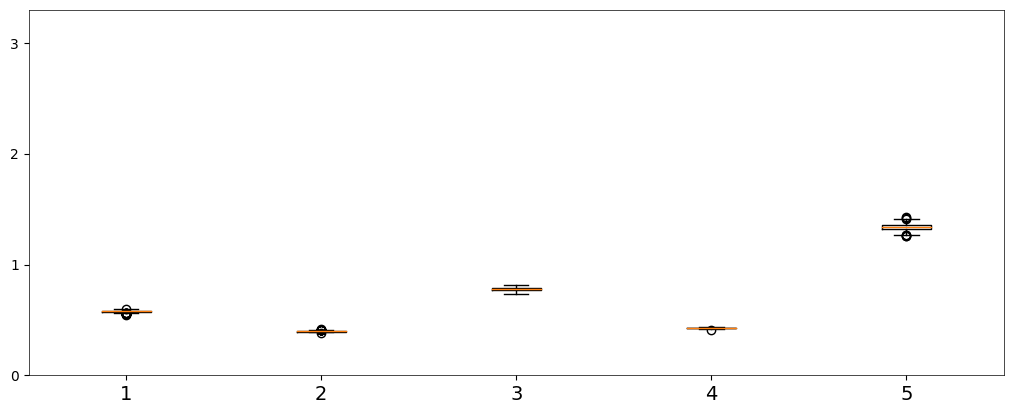}
\caption{$\norm{v_{j}^{h}}$ for \texttt{\_.} (Top) and \texttt{$\langle \text{/s} \rangle$} (Bottom) for the best alignment head over 5 random seeds.}
\label{fig:eos_punct_v_norms}
\end{figure}

\paragraph{Model behaviour.} From the decoder layer depicted in Figure \ref{fig:enc_dec} we can observe that the output of the encoder-decoder attention module is added to the target prefix representation by means of the residual connection $\bm{attn}_{t} + \bm{s}_{t}^{l-1}$. Therefore, the amount of information arriving from the input sequence is determined by the weighted sum of the values. If we analyze the norms of the values vectors (Figure \ref{fig:attn_bypass}) we can see that the source finalizing tokens, especially  \texttt{$\langle \text{/s} \rangle$}, get almost zero norms. This can be interpreted as when assigning high attention weights to these tokens, the Residual + Normalization layer gets almost no information from the source. From the results obtained over 5 random seeds (Figure \ref{fig:eos_punct_v_norms}) we can state that the network picks a common token, i.e. \texttt{$\langle \text{/s} \rangle$} or \texttt{\_.} and projects it to a zero vector through $\mathbf{W_{h}^{V}}$. These results support the \cite{clark-etal-2019-bert} hypothesis about the selection of a token as a "no-op" in the attention mechanism (\texttt{[SEP]} token in BERT model).

\begin{figure}[h!]
	\begin{centering}
	\includegraphics[width=0.48\textwidth]{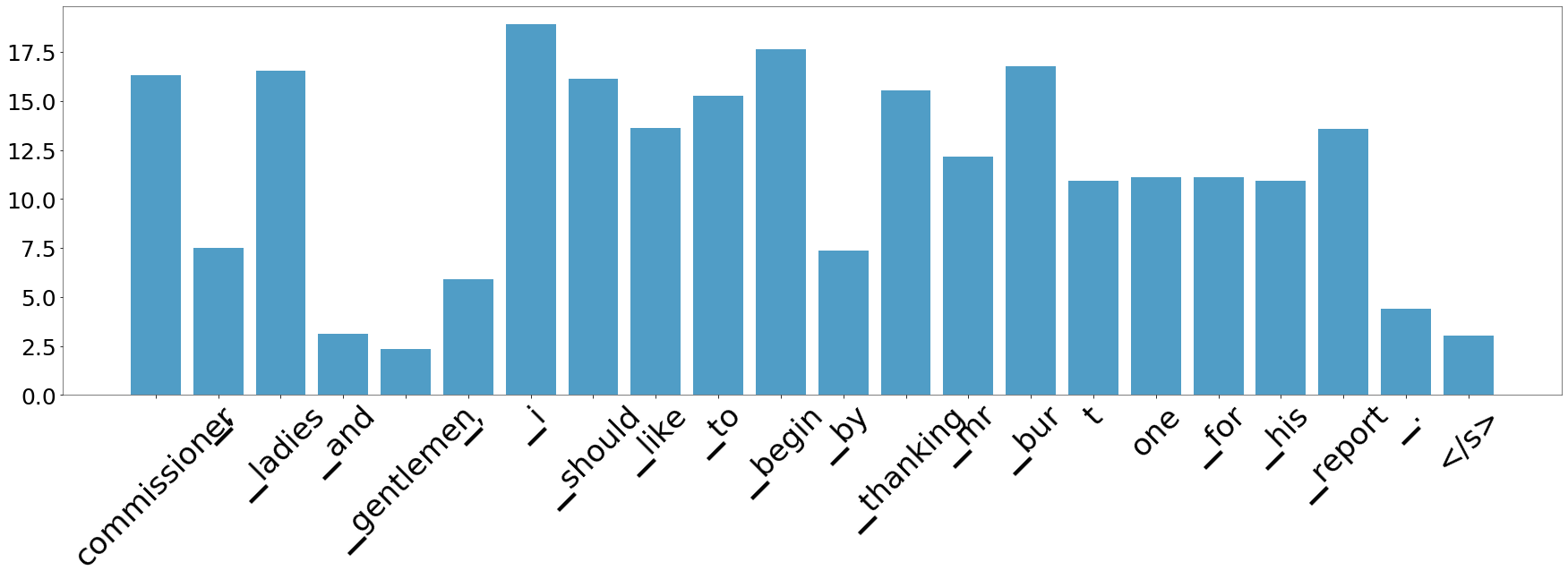}
	\caption{Output representation of the encoder-decoder attention module norms $\norm{\bm{attn}_t}$.}
	\label{fig:attn_norms}
	\end{centering}
\end{figure}

In this way, by putting attention to it, decides how much amount of information flows from the source and target sequences. Note that over the five trained models, the selection of the token that ends up squished varies, for model number 5, the network selects the final punctuation mark as the token used to cancel source contribution. $\bm{attn}_t$ vector norms (Figure \ref{fig:attn_norms}) correlate with our source-target contribution method results depicted in Figure \ref{fig:source_target_contributions}. Representations $\bm{attn}_t$ for tokens such as \texttt{and}, \texttt{gentlemen} and \texttt{\_,} have low norms due to the effect of large attention weights towards finalizing tokens.

\begin{figure}[H]
	\begin{centering}
	\includegraphics[width=0.4\textwidth]{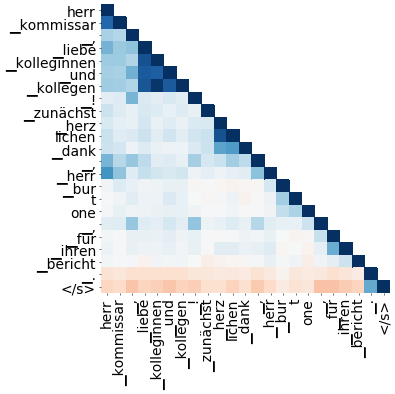}
	\caption{Cosine similarity between encoder representations. Positive similarity (blue), negative (red).}
	\label{fig:cos_encodings}
	\end{centering}
\end{figure}

\begin{figure}[h!]
	\begin{centering}
	\includegraphics[width=0.48\textwidth]{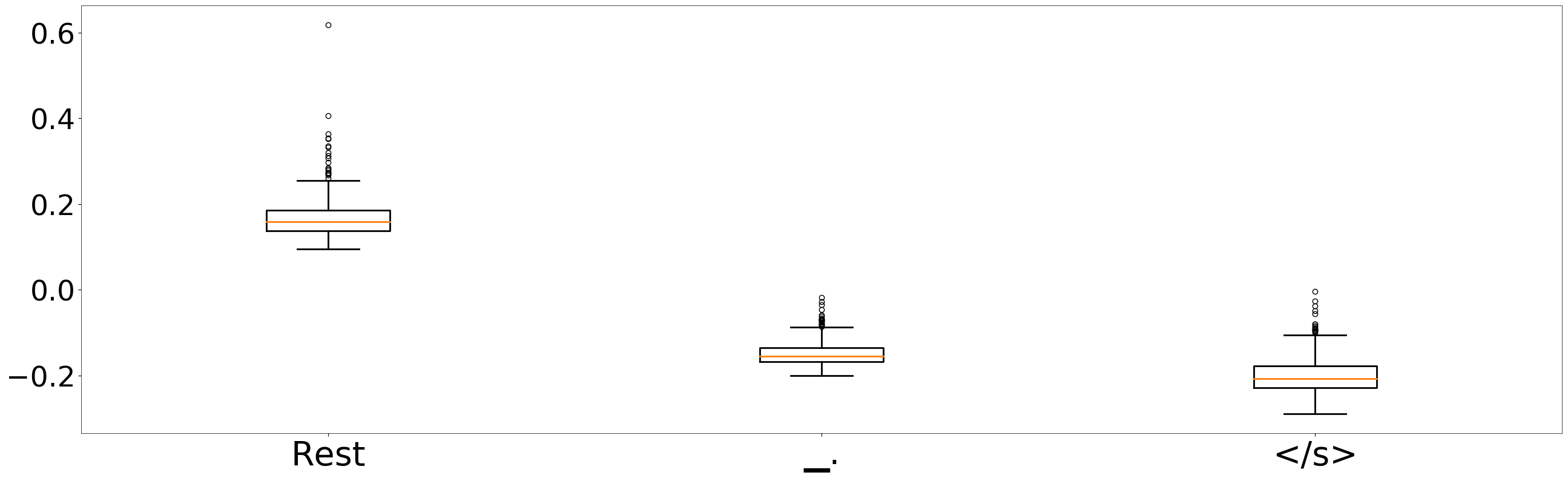}
	\caption{Cosine similarity between encoder representations by type of token across the test set.}
	\label{fig:cosinus_test_set}
	\end{centering}
\end{figure}

Interestingly, the finalizing tokens representations from the last encoder layer show clear differences with respect to the other tokens' representations. Measuring the cosine similarity between every encoder output representation (Figures \ref{fig:cos_encodings}  and \ref{fig:cosinus_test_set}) we observe how the finalizing tokens similarity with every other encoder representation is consistently negative.

We conjecture that these tokens encode minimum information about the source sentence, and the decoder finds them useful in the encoder-decoder attention module to skip source attention. We leave as future work a deeper investigation of this phenomenon.

\subsection{Word Alignment Errors associated to Part-of-Speech}
\label{sec:pos}

If we analyze the percentage of tokens across the whole dataset that are aligned towards finalizing tokens, i.e receiving attention scores greater than $0.5$, we observe (Table \ref{tab:POS_pct_appearance}) that words with a high degree of dependency on the context such as adpositions (ADP), particles (PART) and conjunctions (SCONJ, CCONJ) are likely to get aligned to finalizing tokens. On the other hand, numerical values (NUM), determiners (DET) and verbs (VERB, AUX), which are more independent of the context tend to align to source tokens. We see that functional words are more prone to get aligned to finalizing tokens.

\begin{table}[h!]
\centering\small
\begin{tabular}{cc}
\hline
\textbf{POS-tag} & \textbf{\%} \\ \hline
ADP                                                                           & 49.1                    \\
PART                                                                          & 33.9                    \\
SCONJ                                                                         & 30.7                    \\
NOUN                                                                          & 24                      \\
CCONJ                                                                         & 23.8                    \\
ADV                                                                           & 17.3                    \\
PROPN                                                                         & 16                      \\
PRON                                                                          & 14                      \\
ADJ                                                                           & 13.1                    \\
VERB                                                                          & 12.4                    \\
DET                                                                           & 9.4                     \\
NUM                                                                           & 6.8                     \\
AUX                                                                           & 4.5                     \\
\hline
\end{tabular}
\caption{Words (in \%) aligning to finalizing tokens.}
\label{tab:POS_pct_appearance}
\end{table}

These results agree with our previous observations. Words with a high contribution from the target prefix get attention weights assigned to source finalizing tokens. These results demonstrate a correlation between the attention towards finalizing tokens and the lack of contribution from the source to the model prediction.

\section{Methods to Improve Alignment} \label{sec:improving_alignment}
As shown in the previous analysis, alignments from attention weight matrices reveal errors mainly due to the existence of the skip source attention operation. In this section we propose two methods to get more clear alignments.

\subsection{Heads Importance} \label{sec:heads_importance}
Each layer attention weight matrix is computed by averaging over every head (\S \ref{sec:attn_alignments}). However, we know specific heads learn better alignments \cite{kobayashi-etal-2020-attention}. Based on the Hidden Token Attribution method
\cite{Brunner2020on} we measure the contribution of each head to the output of the model and detect that specialized heads tend to obtain higher contributions. We propose to optimize the extraction of per layer attention weights substituting the naive average approach by a weighted average based on each head contribution. For each head $h$ we compute the summation of the gradients w.r.t the input vectors $\bm{v}^{h}_{j}$:
\begin{align*}
c_{h}(y_{t}) = \sum_{j=1}^{|\textbf{x}|}\norm{\nabla_{\bm{v}^{h}_{j}} P(y_t|\mathbf{y}_{<t},\mathbf{x})}
\end{align*}

Then, to extract its relative contribution, we normalize between the scores of every head:
\begin{align*}
C_{h}(y_{t}) = \frac{c_{h}(y_{t})}{\sum_{h=1}^{H}c_{h}(y_{t})} 
\end{align*}

Finally, we extract hard alignment as a weighted average of the head's relative contribution:
\[
  \mathbf{A}_{t,j} = \left\{
     \begin{array}{@{}l@{\thinspace}l}
       1 \quad j= \argmax_{j'}\sum_{h=1}^{H} C_{h}(y_{t}) \alpha_{t,j'}^{h} \\
       0 \quad \text{else}\\
     \end{array}
   \right.
\]

\subsection{Masking Finalizing Tokens}

Attention shifting towards source finalizing tokens make the models underperform in the Word Alignment Task. From our previous analysis we also demonstrate they are used to manage the amount of information from the prefix that flow to upper layers. We propose to mask the attention weights to the finalizing tokens with zeros to measure the degree of success of secondary attention weights induced alignments.

\subsection{Modified Alignments Results}\label{sec:improving_alignment_results}
Results in Table \ref{tab:aer_results} reflect the reduction in alignment error rate (AER) by applying the proposed methods. Regarding the heads importance method, it improves AER percentage in $2.7$ points in the decoder input ($\mathbf{A}_{i,j}$) alignment setting, although maintaining same accuracy in the decoder output ($\mathbf{A}_{t,j}$) alignments. The difference in improvements in $\mathbf{A}_{i,j}$ are explained by the fact that initial layers have attention heads more specialized, while in the last layers they perform more uniformly. Masking methods reduces $6.3$ and $8.4$ AER points in $\mathbf{A}_{i,j}$ and $\mathbf{A}_{t,j}$ respectively, which indicates that, despite deciding that the prefix contributes the most, the model still pays attention to relevant source tokens.

\begin{table}[h!]
\resizebox{0.49\textwidth}{!}{%
\begin{tabular}{lcccc}
\hline
\multicolumn{1}{c}{\textbf{Method}}  & \textbf{AER} & \multicolumn{1}{l}{\textbf{$\pm$ SD}} & \textbf{AER}    & \textbf{$\pm$ SD}    \\ \hline
                                     & \multicolumn{2}{c}{$\mathbf{A}_{i,j}$}               & \multicolumn{2}{c}{$\mathbf{A}_{t,j}$} \\ \hline
\textbf{Attention weights}           &              &                                       &                 &                      \\
\cite{kobayashi-etal-2020-attention} & 29.8         & 3.7                                   & 47.7            & 1.7                  \\
Ours (HI)                            & 27.1         & 2.0                                   & 47.6            & 1.6                  \\
Ours (Mask)                          & 23.5         & 1.1                                   & 39.3            & 1.5                  \\
Ours (HI + Mask)                     & 22.1         & 1.2                                   & 38.5            & 1.7                  \\
\cite{chen-etal-2020-accurate}       & 20.9         & -                                     & -               & -                    \\ \hline
\textbf{Vector-Norms}                &              &                                       &                 &                      \\
\cite{kobayashi-etal-2020-attention} & 25.0         & 1.5                                   & 41.4            & 1.4                  \\ \hline
\textbf{Word Aligner}                &              &                                       &                 &                      \\
Fast-Align                           & 28.4         & -                                     & 28.4            & -                    \\
GIZA++                               & 21.0         & -                                     & 21.0            & -                    \\ \hline
\end{tabular}
}
\caption{AER results comparison. Our methods are applied on \cite{kobayashi-etal-2020-attention} implementation, which we use as the reference. HI refers to the Heads Importance method (\S\ref{sec:heads_importance}). GIZA++ and Fast-Align results from \cite{Zenkel_2019}.}
\label{tab:aer_results}
\end{table}

\section{Conclusion}
In this paper, we have studied the use of attention weights as an explanatory method for the Transformer in NMT. We have proposed analysis methods that measure the relative contribution of the source and the target prefix sequences. Then, we have demonstrated that the alignment bias towards finalizing tokens, which is the most common alignment error, is used by the model to avoid source information flowing through the decoder. In these cases, the predicted output relies on prefix dependencies, which are identifiable by extending the gradient-based analysis to extract saliency scores. Furthermore, we have proposed two methods to improve the extraction of alignments from attention weights. As future work, we plan to extend our study to more languages pairs, as well as to the multilingual NMT setting.

\section*{Acknowledgements}
We would like to thank Ioannis Tsiamas for the help in the busy days, as well as Carlos Escolano and Christine Basta for the useful comments.
This work is supported by the European Research Council (ERC) under the European Union’s Horizon 2020 research and innovation programme (grant agreement No. 947657).

\bibliography{anthology,custom}
\bibliographystyle{acl_natbib}

\end{document}